\documentclass[letterpaper, 10 pt, conference]{ieeeconf}

\IEEEoverridecommandlockouts                             
\usepackage{cite}
\usepackage{epsfig} 
\usepackage{epstopdf}
\usepackage{amsmath}
\usepackage{bm}
\usepackage{soul}
\usepackage{amsfonts}
\usepackage{color}
\usepackage{listings}
\usepackage{subfigure}
\usepackage{lipsum}

\usepackage{graphicx}
\usepackage{wrapfig}
\usepackage{lscape}
\usepackage{rotating}
\usepackage{dirtytalk}
\usepackage{xcolor,colortbl}
\definecolor{Gray}{gray}{0.9}
\usepackage{url}
\usepackage{hyperref}
\usepackage[makeroom]{cancel}
\usepackage{pifont}

\usepackage{multirow}

\usepackage{graphicx}
\usepackage{algorithm}
\usepackage{algorithmic}

\usepackage{mdframed}

\title{\large \textbf{Design and Validation of a Multi-Arm Relocatable Manipulator for Space Applications}}

\author{Enrico Mingo Hoffman$^{1, 3}$, Arturo Laurenzi$^{1}$, Francesco Ruscelli$^{1}$, Luca Rossini$^{1}$, \\
Lorenzo Baccelliere$^{1}$, Davide Antonucci$^{1}$, Alessio Margan$^{1}$, Paolo Guria$^{1}$,  Marco Migliorini$^{1}$, \\
Stefano Cordasco$^{1}$, Gennaro Raiola$^{1, 3}$, Luca Muratore$^{1}$, Joaquín Estremera Rodrigo$^{2}$, \\
Andrea Rusconi$^{3}$, Guido Sangiovanni$^{3}$ and Nikos G. Tsagarakis$^{1}$
\thanks{$^{1}$Humanoids \& Human Centred Mechatronics Lab., Istituto Italiano di Tecnologia (IIT), Via Morego 30, 16163 Genova, Italy, {\tt\small nikos.tsagarakis@iit.it} \newline
$^{2}$GMV, Isaac Newton 11 Tres Cantos 28760, Madrid, Spain, {\tt\small jestremera@gmv.com} \newline
$^{3}$Leonardo S.p.A., Viale Europa, 20014, Nerviano, Italy, {\tt\small andrea.rusconi@leonardocompany.com, guido.sangiovanni@leonardocompany.com}
\newline
The development of MARM is funded by the Europen Space Agency (ESA) through the MIRROR project.}}

\begin{document}

\maketitle

\begin{abstract}
This work presents the computational design and validation of the Multi-Arm Relocatable Manipulator (MARM), a three-limb robot for space applications, with particular reference to the MIRROR (i.e., the Multi-arm Installation Robot for Readying ORUs and Reflectors) use-case scenario as proposed by the European Space Agency.

A holistic computational design and validation pipeline is proposed, with the aim of comparing different limb designs, as well as ensuring that valid limb candidates enable MARM to perform the complex loco-manipulation tasks required.
Motivated by the task complexity in terms of kinematic reachability, (self)-collision avoidance, contact wrench limits, and motor torque limits affecting Earth experiments, this work leverages on multiple state-of-art planning and control approaches to aid the robot design and validation. These include sampling-based planning on manifolds, non-linear trajectory optimization, and quadratic programs for inverse dynamics computations with constraints.
Finally, we present the attained MARM design and conduct preliminary tests for hardware validation through a set of lab experiments.

\end{abstract}

\IEEEpeerreviewmaketitle

\section{Introduction and Related Works}\label{sec:introduction}

Space applications present unique conditions and opportunities for the development of robotics platforms: although robotics technologies for terrestrial locomotion and manipulation are rather advanced, conditions in the orbital environment, such as microgravity, pose special requirements for the design and control of such robots.
The  International Space Station (ISS) is a modular space station that serves research purposes conducted in different science and engineering fields, and as a human outpost in low Earth orbit for long-duration missions. 
In order to assist humans in the various activities on the station, several robotic systems have been developed and deployed.
Internal and external robotics in low earth orbits encompass a variety of robotic platforms to facilitate the inspection, maintenance, assembly, and repair of the station. 
This includes both humanoids such as the NASA Robonaut \cite{ambrose2000robonaut} to assist and work alongside the astronauts and robotic arms mounted outside the station such as the Canadarm2 \cite{hiltz2001canadarm}, the Japanese Experiment Module Remote Manipulator System (JEM-RMS) \cite{matsueda1991jemrms} and the European Robotic Arm (ERA) \cite{boumans1998european} which are instead used to carry out extravehicular activities (EVAs).
\par
In this context, we present MARM, a Multi-Arm Relocatable Manipulator for loco-manipulation in microgravity capable of self-relocate through the station to perform a variety of tasks, designed and manufactured by the Humanoid and Human Centered Mechatronics (HHCM) Lab, at Istituto Italiano di Tecnologia (IIT), in collaboration with Leonardo S.p.A and GMV.
The robotic platform was designed following a computational design (co-design) approach to analyze and optimize two critical criteria: platform mobility and flexibility. 
The devised robot consists of a central body that connects three limbs with a latching end-effector, through which power and data are fed to and from the robot. 
Besides the transmission, the robot can use the arms to grapple standard interconnects (SIs), installed on re-configurable tiles over the station's surfaces for locomotion purposes, tile assembly, and handling of Orbital Replacement Units (ORUs).
In order to provide mechanical, data, and power connections to the robot during the execution of the activities, the SI choice for this project is the Standard Interface for Robotic Manipulation (SIROM) \cite{vinals2020standard}.
The SIROM provides a multi-functional interconnect that combines mechanical, electrical, data, and fluid interfaces into a single connection, thus simplifying the design of the robot end-effector.
The grasped payloads can be relocated from one position to another, hence the robot must have the manipulation ability to retrieve the payload from the body storage or surface, travel holding the transported payload, and assemble it in the desired location.
%
\begin{figure}[t!]
    \centering
    \includegraphics[width=0.95\columnwidth]{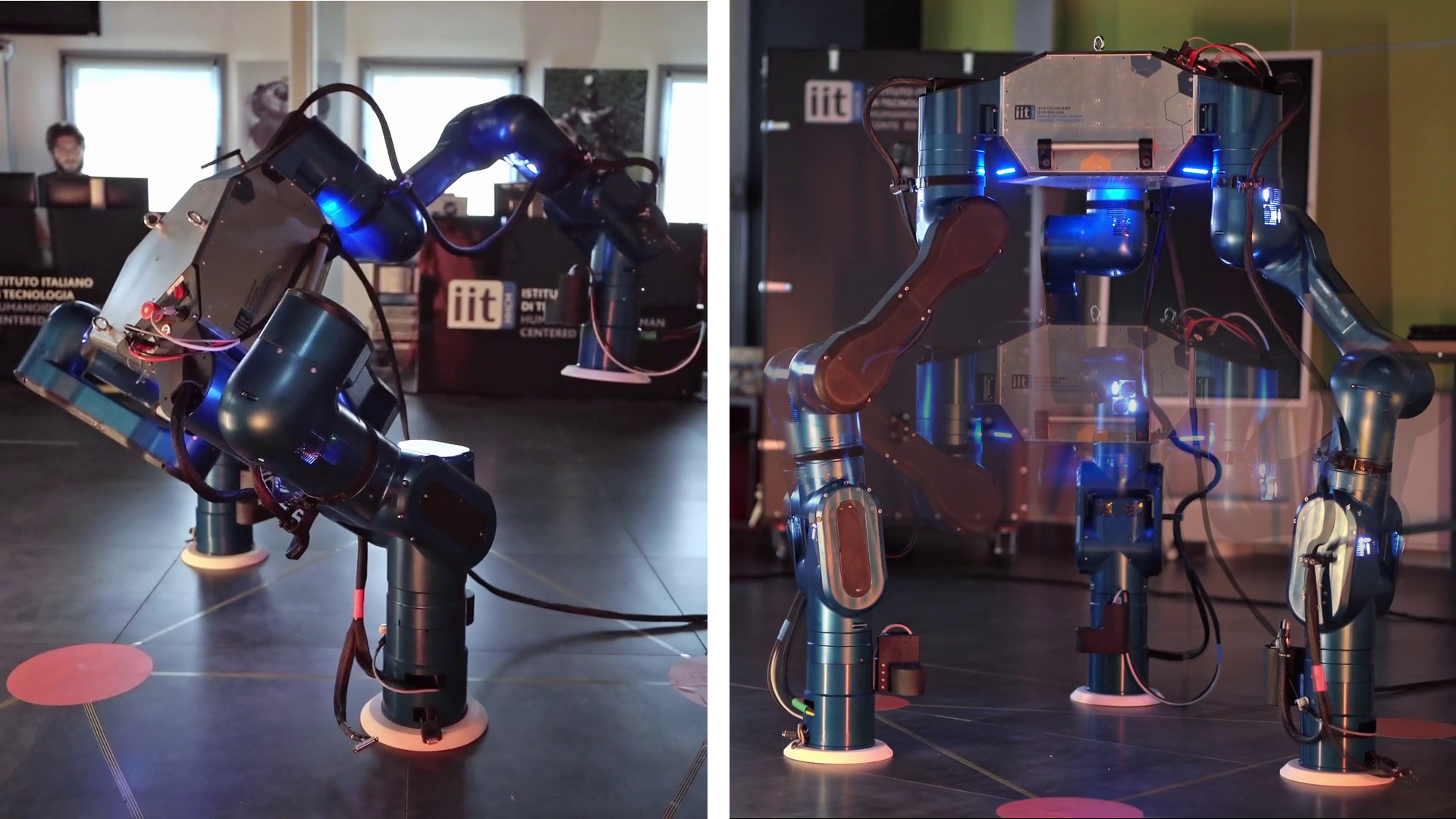}
    \caption{Final MARM design carrying out Cartesian space validation tests. Double stance (left), and squatting motions (right).}
    \label{fig:mirror_video_snaps}
    \vspace{-0.5cm}
\end{figure}
In particular, the design and development of MARM are adapted for the assembly of hexagonal prism tile with a dimension of $1.2 \ m$ (side to side), thickness $0.2 \ m$, weighting around $12 \ kg$, to form a larger structure. 
This application - named \textit{MIRROR} (Multi-arm Installation Robot for Readying ORUs and Reflectors) - is pursued by the European Space Agency (ESA). 
The multi-arm design and the dexterity of the MARM robot make the platform suitable for different types of space applications such as manufacturing and assembly \cite{roa2017robotic}, servicing utilizing in-situ resources \cite{mukherjee2020robotically,lee2016architecture}, berthing and catching of satellites \cite{yoshida2006capture}.
Similarly to Canadarm2/ERA robots, MARM should be able to relocate itself on the space station by crawling. 
The main difference between our system and the aforementioned platforms is the presence of extra limbs attached to a central base which can serve as a payload bay to facilitate the manipulation and assembly of large payloads \cite{papadopoulos1993large}. 
This difference allows the robot to perform full body motions using the kinematics of the floating base. 
Furthermore, once in position, the arm(s) used for anchoring can facilitate the assembly operation by adjusting the MARM central base, reducing the forces transferred to the station  truss. 
This is one of the main advantages of having multiple arms instead of using the same arm for crawling and assembly as indicated in \cite{mcbryan2020comparison}.
Moreover, the presence of multiple arms allows the robot to perform bi-manual tasks which a robot like the Canadarm2 can only perform with the Dextre robotic platform mounted as end-effector \cite{rusconi2008dexarm}.
\par
The first phase of the design and development of the MARM focused on the system specifications of the robot and limb architecture given the expected mechanical loads and actuation requirements.
This phase has been carried out within a co-design cycle using a simulated robot prototype. This allowed the characterization of the kinematics that enables to reduce the weight and favor the compactness by determining the essential Degrees of Freedom (DoFs).
The second phase consisted of an in-depth analysis of reachability, manipulation, (self-)collision avoidance, singularities, and joint torque, as well as contact wrenches, with the aim of evaluating the performance of the robot design.
\par
The paper is organized as follows. Section~\ref{sec:codesign} presents the proposed co-design and validation pipeline; this pipeline is employed to generate the final design and of the MARM prototype, that is presented in Section~\ref{sec:proto}. Preliminary tests and experiments using the real platform are presented in Section~\ref{sec:experiments}. Finally, Section~\ref{sec:conclusions} concludes the work with a summary of the achieved results, and future work directions.

\section{Computational Design}\label{sec:codesign}
The preliminary design studies of the MARM robot has been driven by the following guidelines:
\begin{itemize}
    \item symmetry: to simplify planning, control, and maintenance;
    \item modularity: by re-using the same actuator modules;
    \item compactness: to save weight and space by reducing the number of DoFs;
    \item multipurpose kinematics: suitable for generic loco-manipulation tasks.
\end{itemize}
These design guidelines were steered by three concrete task scenarios that the proposed robot will be able to carry out.
According to such scenarios, each of the designed robot end-effectors shall have the ability to (i) carry a tile, (ii) to be used as a stance foot locked through a SIROM mechanical interface to the ground~\cite{vinals2020standard}, and (iii) to hold a visual probe that provides near-field perception during manipulation and assembly phases. 
\par
To derive the kinematic architecture of the limbs of the MARM robot, a series of tests, simulations, and analyses were conducted. These include (i) sampling-based motion planning, (ii) trajectory optimization (TO), and (iii) inverse dynamics computations. 
%
The aim of such studies is to provide a \textit{holistic validation pipeline} to assess if a given design is compatible with the MIRROR tasks, in terms of both reachability as well as  considering the maximum interaction forces that the SIROM interface can resist, which are in the order of $5 \ kN$ for the compression and radial loads, $420 \ Nm$ for the maximum axial torque, and $150 \ Nm$ for the bending moments~\cite{sirom_datasheet}. 
As an additional outcome of the proposed validation pipeline, maximum joint torques to guide the sizing and selection of the MARM actuators are also obtained.
%
\par
\begin{figure}[htb!]
    \vspace{-0.25cm}
    \centering
    \includegraphics[width=1.\columnwidth, trim={2cm 0cm 2cm 0.5cm}, clip=true]{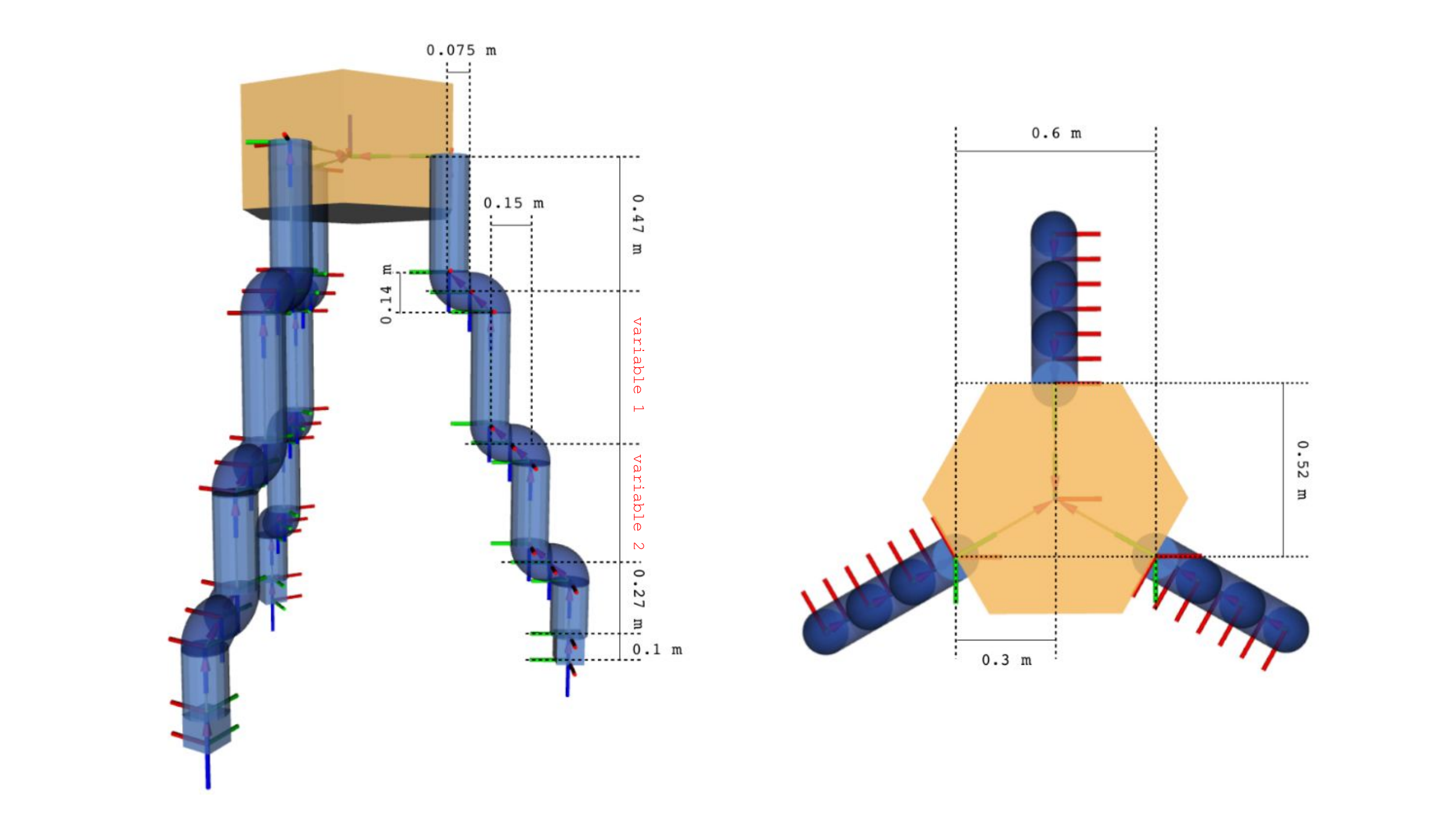}
    \vspace{-0.8cm}
    \caption{Overview of the preliminary MARM breadboard kinematics and dimensions considered.}
    \label{fig:initial_guess}
\end{figure}

As initial candidates for the kinematics structure of the MARM limb, we considered 6 and 7-DoFs anthropomorphic arms, mounted orthogonally to the robot's central base in a symmetric fashion. 
With the aim to maximize the modularity of the resulting limb, we construct its kinematics by repetition of identical yaw and pitch modules (see Figure~\ref{fig:initial_guess}). 
In addition, we also consider an in-line ankle assembly made with a customized pitch-yaw 2-DoFs module. 
Minimal link lengths have been considered, which depend on an estimate of the actuators' bulk, which was conservatively based on the dimensions of the most powerful actuator model available. 
The resulting total mass is in the order of $90 \ kg$: $25 \ kg$ per limb, plus a $15 \ kg$ base link.
The set of tunable kinematic parameters is therefore given by the two main link lengths, as illustrated in Figure~\ref{fig:initial_guess}.
%
\par
The aim of the robot co-design is the selection of the best parameters combination that allows the kinematic task execution while respecting the geometric and kino-dynamic constraints, according to the simulation previously mentioned. 
This goal has been accomplished through a streamlined pipeline that programmatically generates a robot model by iteratively increasing the adjustable link lengths and evaluates its feasibility by attempting to solve the validation pipeline that is described in Section~\ref{subsec:mpid} and Section~\ref{subsec:to}.

\subsection{Motion planning and inverse dynamics}\label{subsec:mpid}
\begin{figure}[htb!]
    \centering
    \includegraphics[width=.9\columnwidth, trim={4cm 1.5cm 4cm 2cm}, clip=true]{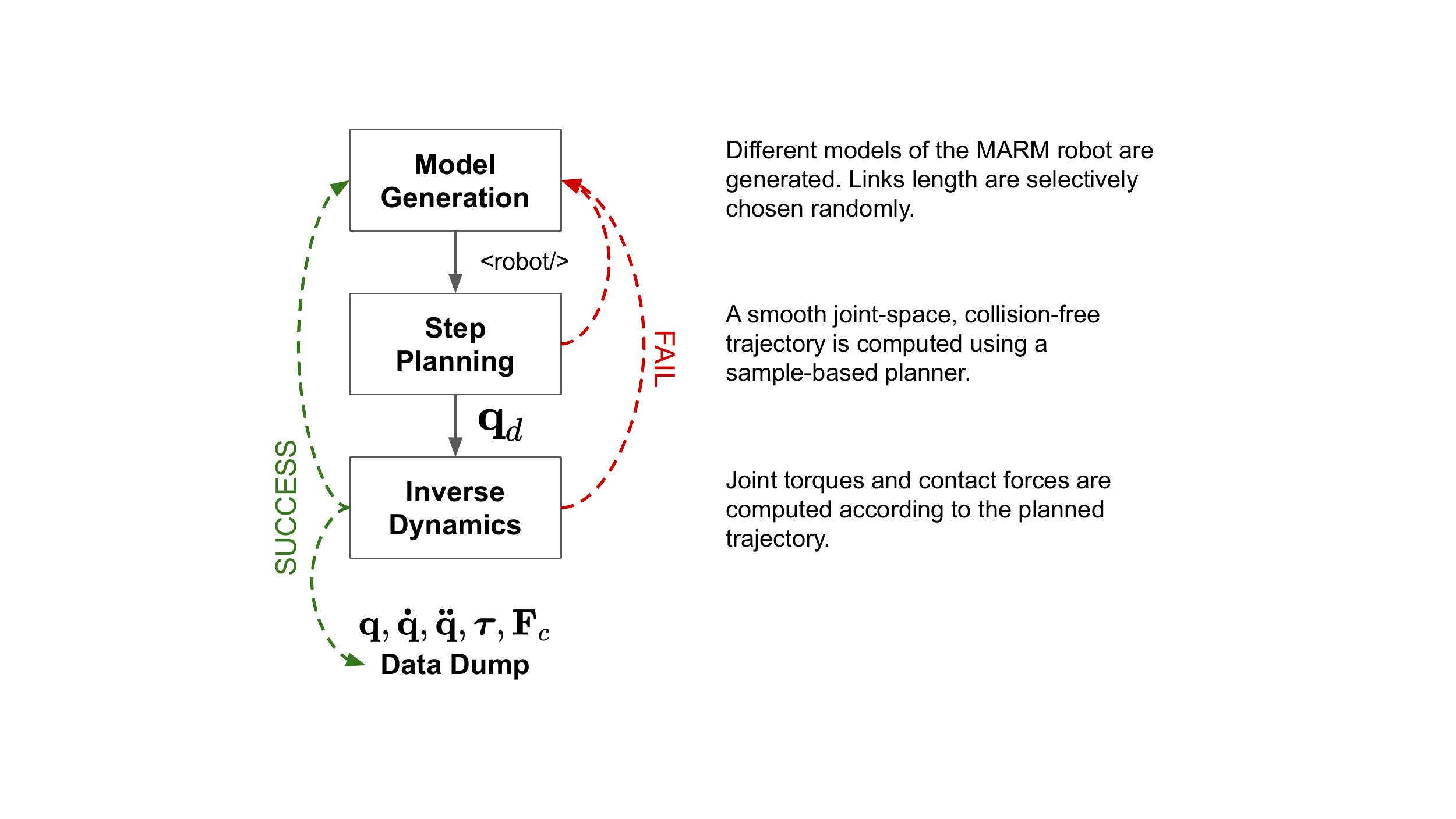}
    \vspace{-0.5cm}
    \caption{Simulation and analysis pipeline.}
    \label{fig:pipeline}
\end{figure}

Each model generated by our pipeline, shown in Figure~\ref{fig:pipeline}, is tested in two types of scenarios in order to evaluate link lengths, joint torques, and contact wrenches both in the presence and absence of gravity:
\begin{itemize}
    \item in double stance configuration while the free limb moves toward the next tile at the given distance of $1.5 \ m$, with and without carrying the tile payload,
    \item in single stance configuration while one of the other limbs moves toward the next tile at the given distance of $1.5 \ m$, with the second free limb carrying a tile.
\end{itemize}
%
%
While the first scenario permits the evaluation of the kinematics reachability in a constrained double stance case, the second one is the most challenging in terms of loading and contact wrenches because a single limb of the MARM robot must support the weight of the entire robot. 
Despite this may not be a particular problem in a space scenario, where the gravitational load is negligible and joint accelerations are predominant, it becomes crucial for Earth experiments.
\par 
We rely on a planning pipeline based on OMPL~\cite{sucan2012ompl} to generate a collision-free trajectory for the MARM candidate that connects the two adjacent stances.
To obtain a contact-consistent motion, the planning problem must be constrained inside the non-linear manifold given by the contacts.
A projection approach has been therefore used to guide the expansion of the search tree through sub-spaces tangent to the manifold itself, guaranteeing constraints consistency.
Finally, feasible start and goal configurations that enhance the reciprocal connectivity between stances have been generated following the work done in \cite{nspg}.
%
%
%
%
%

\begin{figure}[htb!]
    \centering
    \includegraphics[width=1.\columnwidth, trim={3.5cm 2cm 3.5cm 1cm}, clip=true]{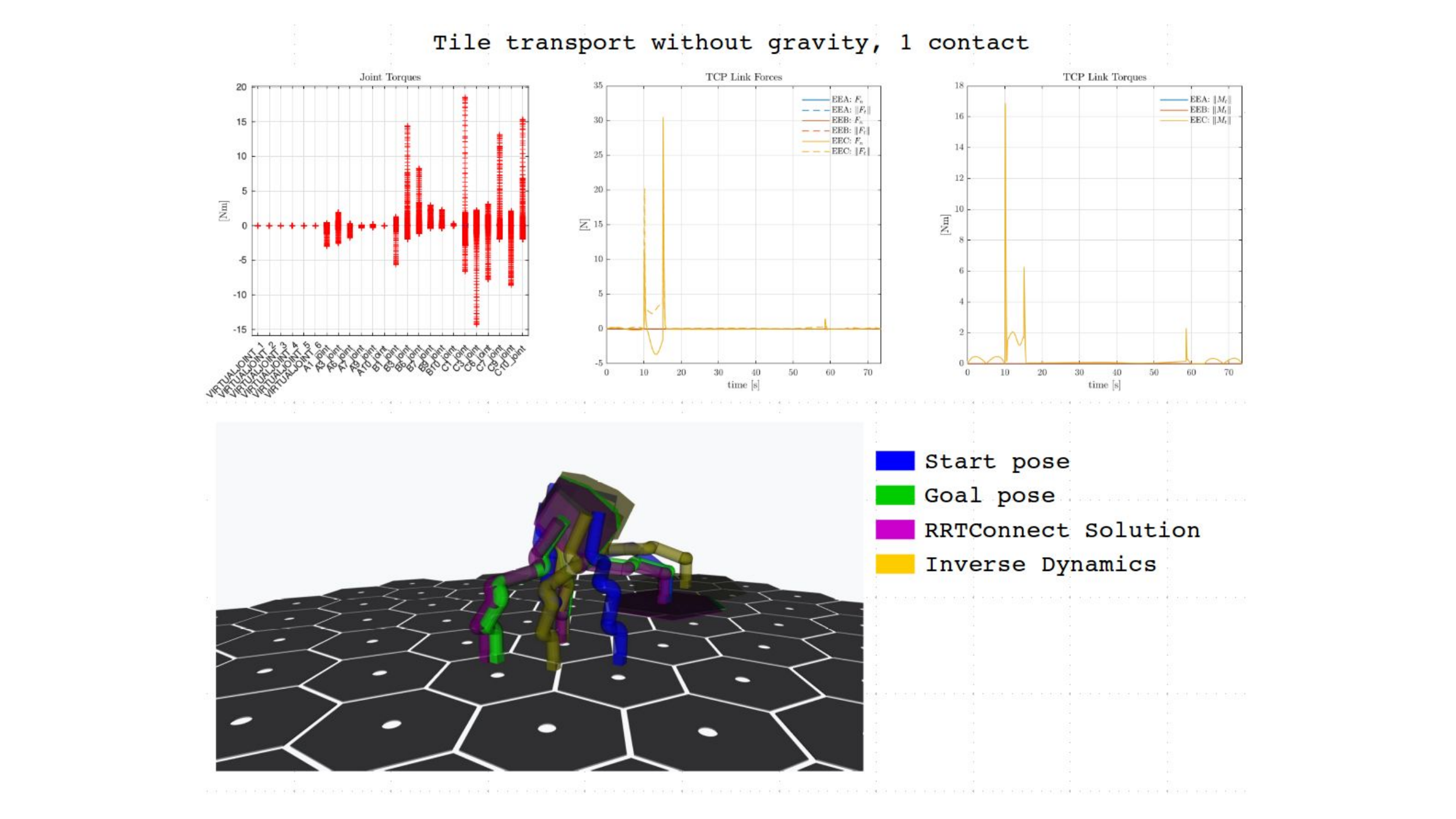}
    \vspace{-0.5cm}
    \caption{Tile transport without gravity, single stance worst case. For every newly generated robot, a start and a goal configuration, in blue and green respectively, are randomly generated lying on the contact manifold. The kinematic, collision-free solution in violet from the sample-based planner is computed connecting the start and the goal. Finally, the inverse dynamics to track the solution is evaluated, in yellow.}
    \label{fig:no_gravity_single_contact_sbp}
\end{figure}

The main objective for this set of simulations is to evaluate the kinematic capability of the generated topology in performing the considered motion scenarios subject to worst-case conditions, i.e. under $1g$ gravity acceleration and with one arm holding the tile payload.
In order to compute the required joint torques and contact forces to perform the planned trajectory, we run an inverse-dynamics based solver to find the joint accelerations and contact forces that realize the reference joint positions given the floating-base dynamics and contact constraints.
The resulting optimization problem can be solved by means of Quadratic Programming optimization~\cite{OpenSot17, laurenzi2018balancing}.
Joint torques can then be computed from the optimal joint accelerations and contact forces, while joint positions and velocities are computed by integrating the obtained joint accelerations.
For the sake of brevity, we here report the results in the case of single stance contact, while carrying a tile with no gravity, for the selected kinematics, see Figure~\ref{fig:no_gravity_single_contact_sbp}.
%
%
%
The top graph on the left present a box plot of the joint torques, which for the no gravity case are very low, under $20 \ Nm$, also for decent fast movements. 
The second and third graphs show the force and torques respectively, at the Tool Center Point (TCP), representing the norm of the interaction forces between the foot of the MARM robot and the SIROM interfaces, that are under the maximum allowed.
These experiments permitted us to assess that the selected kinematics of the MARM robot limb is capable to perform the required motion tasks.

\subsection{Trajectory Optimization for the single stance scenario in presence of gravity} \label{subsec:to}
While in absence of gravity, the motion and carrying strategy are not particularly important for the purpose of minimizing the joint torques and contact forces, it became fundamental in the Earth scenario of the MARM robot where preliminary experiments will be carried on, particularly for the most challenging case of a single limb contact with the ground while the robot is taking the next step or performing a manipulation task with the other limbs.
Clearly, the minimum-effort pose for the stance leg corresponds to the fully stretched configuration of the stance leg.
%
However, this configuration is kinematically infeasible, as it does not allow the swing foot to extend enough to cover its required motion, and a trade-off arises.
\par
\begin{figure}[htb!]
    \centering
    \includegraphics[width=1.\columnwidth, trim={2cm 4cm 2cm 4cm}, clip=true]{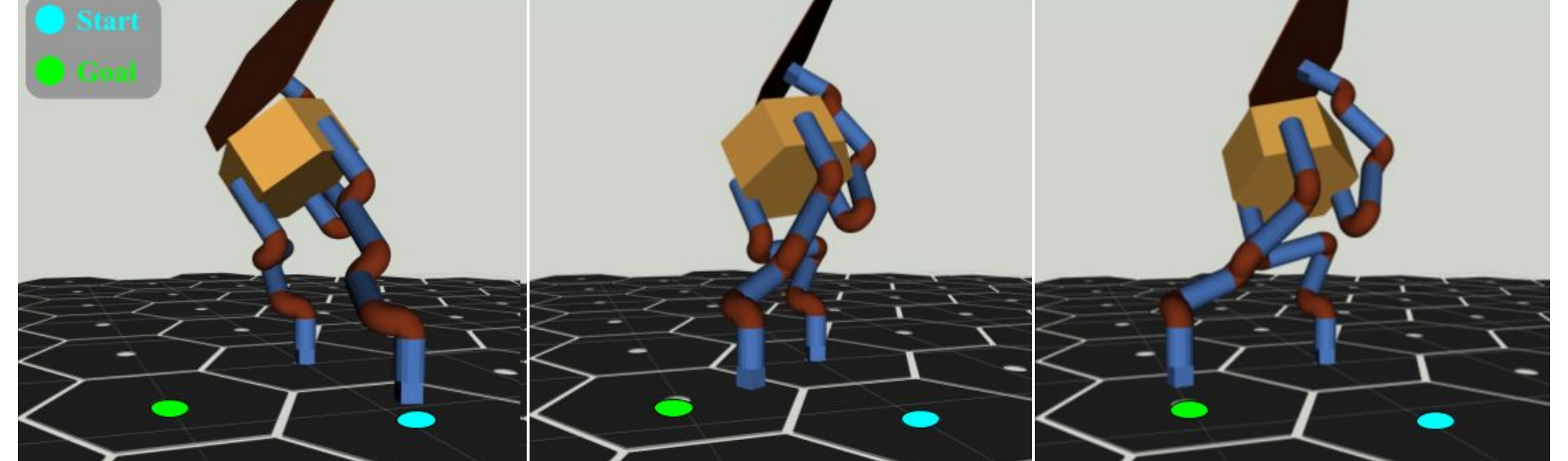}
    \vspace{-0.5cm}
    \caption{Snapshots of the simulation studies executed with the optimized trajectories.}
    \label{fig:to}
\end{figure}
\begin{figure}[htb!]
    \centering
    \includegraphics[width=1.\columnwidth, trim={0cm 12cm 1cm 1cm}, clip=true]{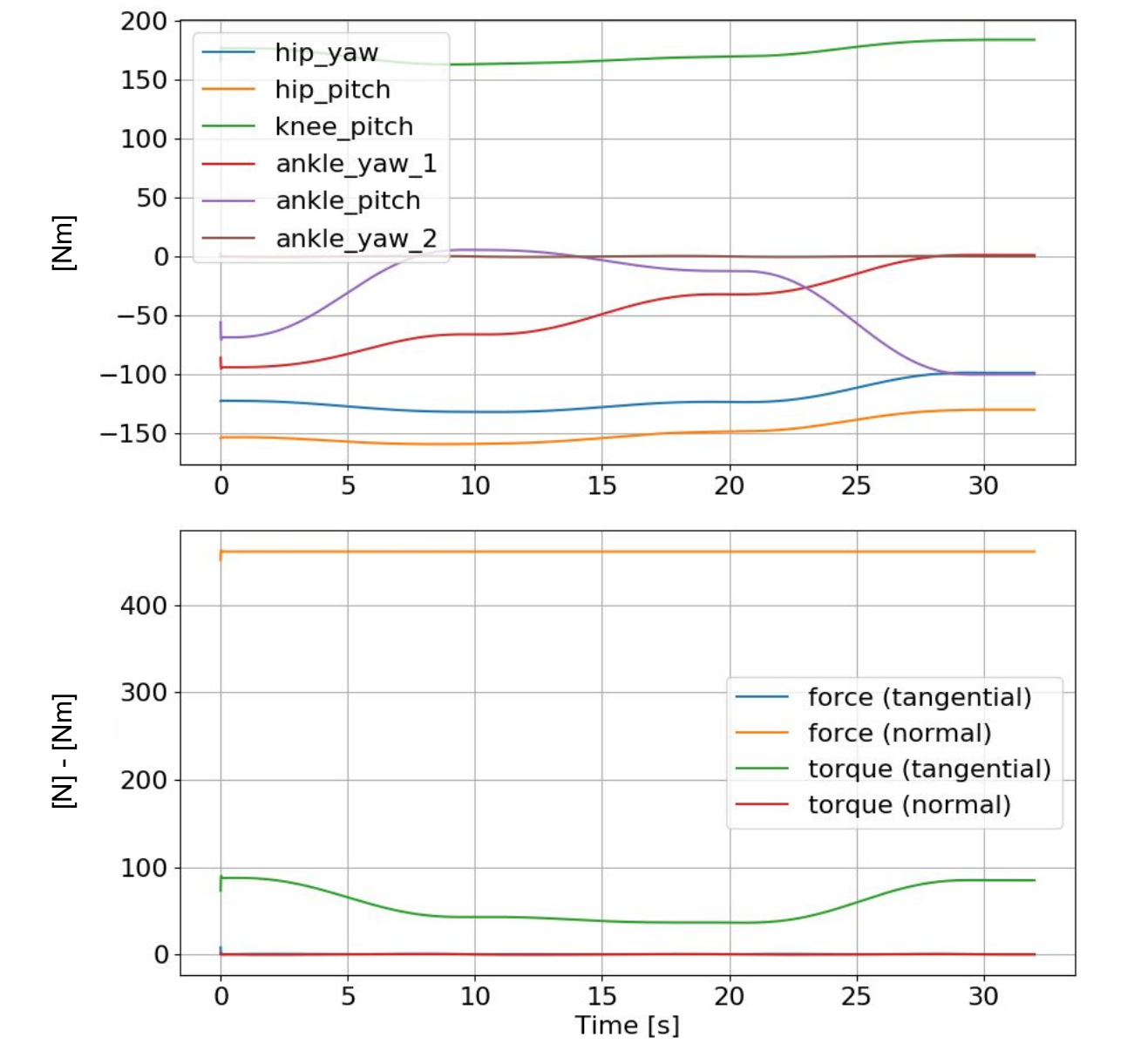}
    \vspace{-0.8cm}
    \caption{Joint and contact mechanical loading with the optimized trajectories for the single foot at stance case under full gravity condition.} \label{fig:to_results}
\end{figure}

In order to compute a pose that is compatible with both the task kinematics and the assumed maximum static load from the motors, we set up the following TO problem where the joint space trajectory is optimized to minimize motion under quasi-static assumption, subject to 
\begin{itemize}
    \item kinematic task requirements (i.e., the MARM candidate must be able to move the swing foot between adjacent SIROM sockets);
    \item torque limits;
    \item contact force limits due to the SIROM interface.
\end{itemize}
On top of these basic tasks, we add further constraints and cost terms to avoid self-collisions, promote a horizontal orientation of the robot base link, and regularize the value of the contact wrench. The resulting non-linear TO problem is formulated and solved using the recently published Horizon~\cite{ruscelli2022horizon} framework.
If feasible, the optimal solution results in the required minimum effort posture, plus an intuitive motion strategy: the robot pivots around its contact point by moving the most distal yaw joint of the stance leg, as shown in Figure~\ref{fig:to}.
To evaluate the total joint torques and contact wrench, including dynamical contributions from the robot inertia, we feed the obtained joint space trajectory into our inverse dynamics solver, yielding the results shown in Figure~\ref{fig:to_results}.
Notice how the hip and knee joint never exceed $200 \ Nm$ torque, whereas ankle joint torques are kept below the $100 \ Nm$ threshold.

%

\subsection{Workspace and reachability analysis}
\begin{figure}[htb!]
    \centering
    \includegraphics[width=1.\columnwidth, trim={5cm 2cm 5cm 7cm}, clip=true]{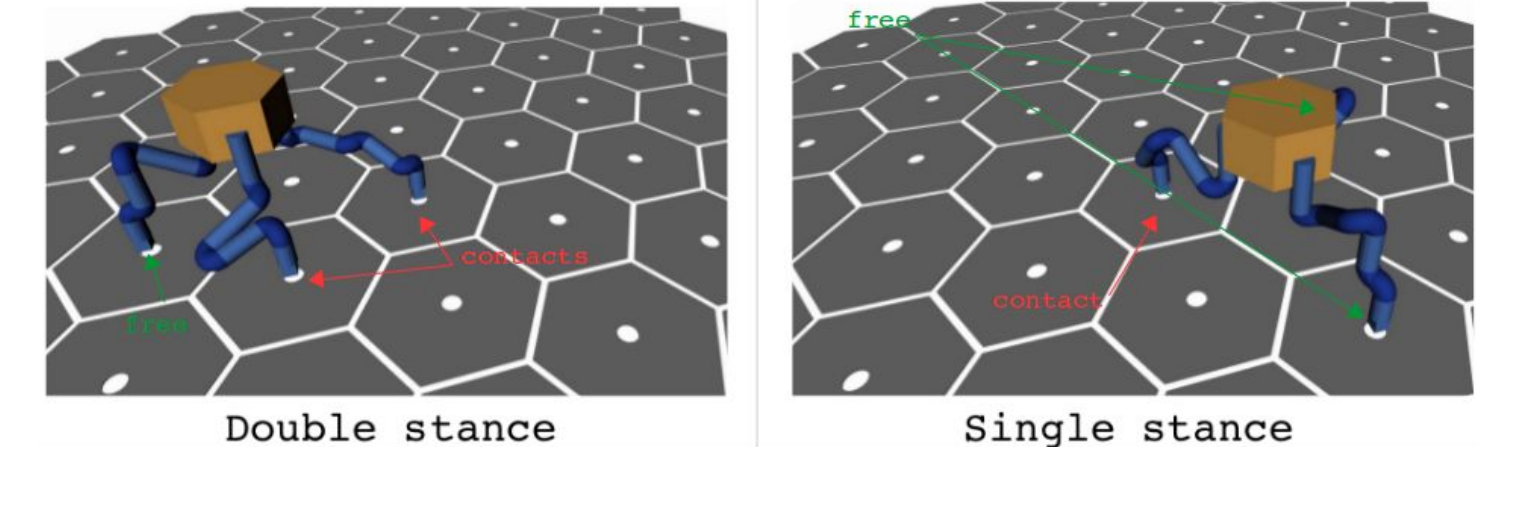}
    \vspace{-1cm}
    \caption{Free limb reachability trial with the robot at double and single stance support.}
    \label{fig:IK}
\end{figure}

In this set of numerical simulations, we tested the capabilities of the selected kinematics and dimensions in reachability tasks. 
We first consider a reachability problem, which we solved using the CartesI/O framework~\cite{laurenzi2019cartesi}. At first, the robot is in double stance and uses the free leg to reach two adjacent tiles, placed on the left and on the right respectively, of the actual tile. 
This represents a very constrained case, however, the robot is capable to move the remaining free leg and the base to the two goals, see Figure~\ref{fig:IK} (left sub-plot).  
Consecutively, one leg is detached from the ground passing from a double stance to a single stance configuration. 
In this case, the robot is capable to reach other tiles placed further with the free leg, see Figure~\ref{fig:IK} (right sub-plot). 
In both cases, we emphasize the possibility to keep the base upright while moving the end-effectors at constant orientation. 
\par
\begin{figure}[htb!]
    \centering
    \includegraphics[width=0.95\columnwidth, trim={3cm 3cm 3cm 3cm}, clip=true]{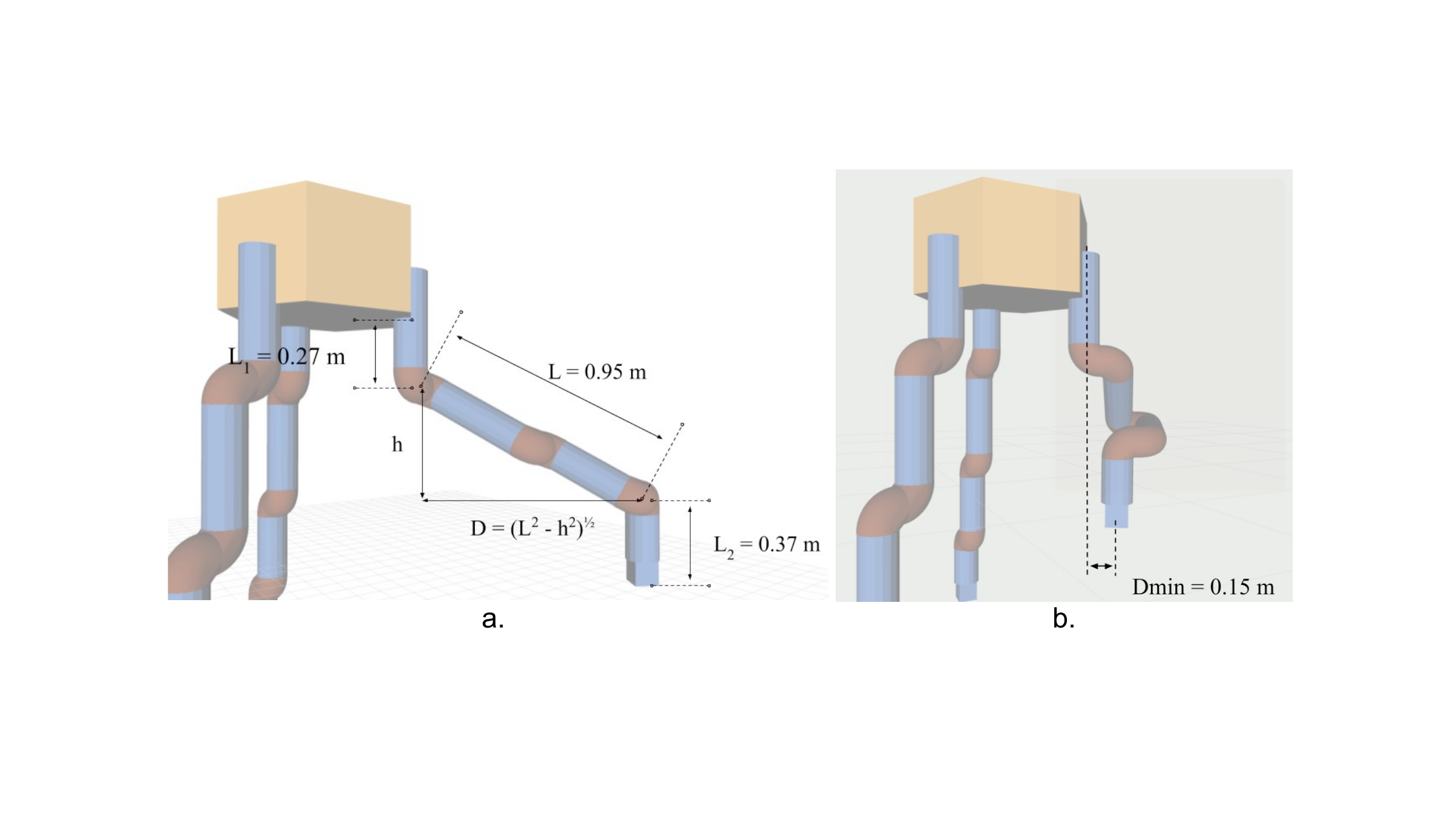}
    \vspace{-0.25cm}
    \caption{On the left, reachability and workspace considerations of the selected limb kinematics. On the right, near workspace boundaries due to the specific kinematic choice.}
    \label{fig:ws}
\end{figure}
\begin{figure}[htb!]
    \centering
    \includegraphics[width=0.8\columnwidth, trim={7.5cm 4cm 6.5cm 4cm}, clip=true]{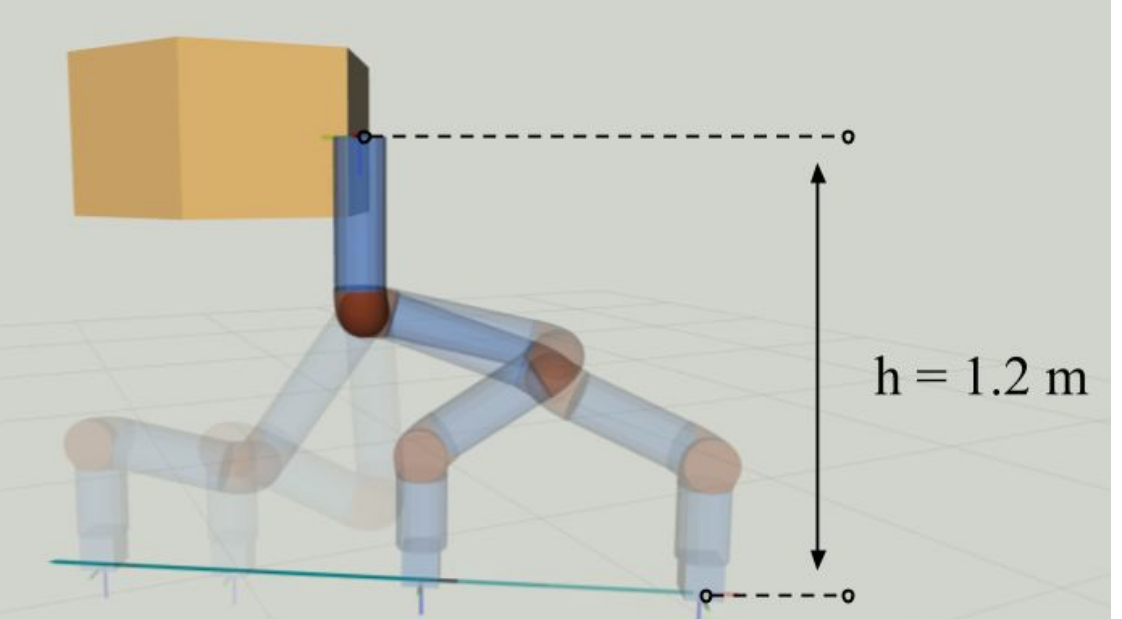}
    \caption{Nominal trajectory path used to compute the manipulability of the kinematic arrangement.}
    \label{fig:manip_trj}
\end{figure}
\begin{figure}[htb!]
    \centering
    \includegraphics[width=0.95\columnwidth, trim={1cm 0.5cm 1cm 0.5cm}, clip=true]{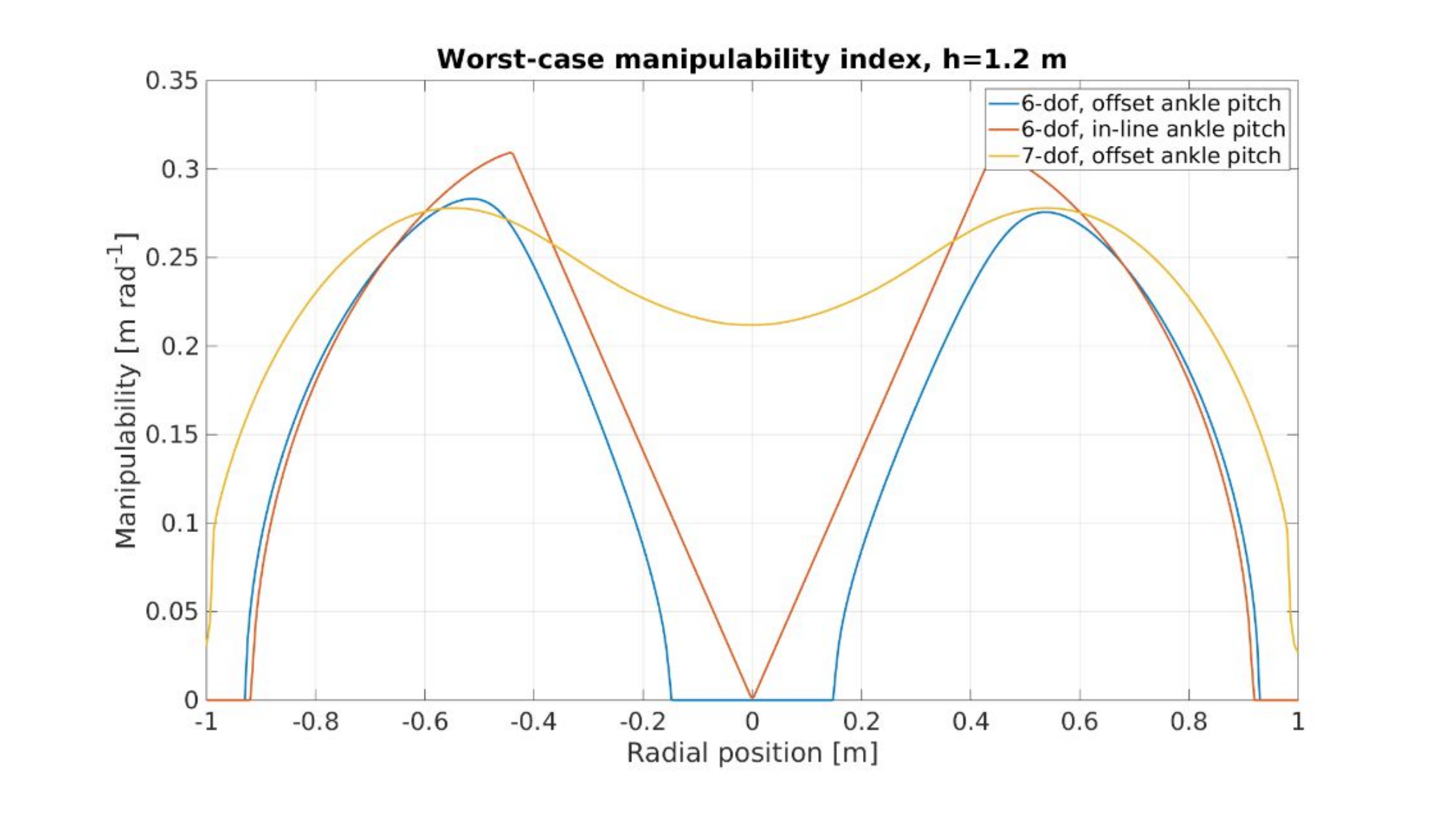}
    \vspace{-0.25cm}
    \caption{Manipulability index along the defined trajectory path of Figure~\ref{fig:manip_trj} for the configuration in the simulation studies for the 6-DoFs with offset ankle, 6-DoFs with in-line ankle, and the 7-DoFs with offset ankle.}
    \label{fig:manip_plot}
    \vspace{-0.5cm}
\end{figure}

Secondly, we analyzed the workspace focusing on one single vertical plane passing through the most proximal yaw motor axis.
The overall workspace in 3D space is then obtained by rotating the obtained planar workspace around the yaw axis itself. Furthermore, is it assumed that the end effector approach axis must be kept perpendicular to the ground plane in order to ensure correct locking to the SIROM socket. 
In this scenario, for any given base height, the far boundary of the workspace is obtained by applying simple trigonometric rules, as shown in Figure~\ref{fig:ws}a.
The near workspace boundary is instead affected by the specific kinematic arrangement of choice. In particular, selecting pitch joints with offset input/output links allows for an optimal folded configuration, but restricts the inner workspace so that $D \geq D_{min}$, with $D_{min}$ being equal to the pitch offset length, as shown in Figure~\ref{fig:ws}b.
Such a restriction can be avoided:
\begin{itemize}
    \item by adopting an in-line ankle pitch joint, at the expense of optimal foldability;
    \item by adding a yaw joint after the first pitch joint, increasing the system complexity and total weight and leading to a 7-DoFs arrangement.
\end{itemize}
Such solutions can be compared in terms of worst-case manipulability, that is the ratio between the end-effector velocity along the least controllable direction and the norm of the joint velocity vector that achieves it.
Computing such an index over the nominal trajectory shown in Figure~\ref{fig:manip_trj}, results in the plot shown in Figure~\ref{fig:manip_plot}. 
It can be noticed how the workspace near-limit of the 6-DoFs offset ankle design (blue line) is reduced to a single point with an in-line ankle design (red line). 
Furthermore, as expected, a 7-DoFs offset ankle design completely removes any singularity point inside the robot arm workspace (yellow line).

\section{Final Design and Prototype}\label{sec:proto}
\begin{figure}[htb!]
    \centering
    \includegraphics[width=.95\columnwidth]{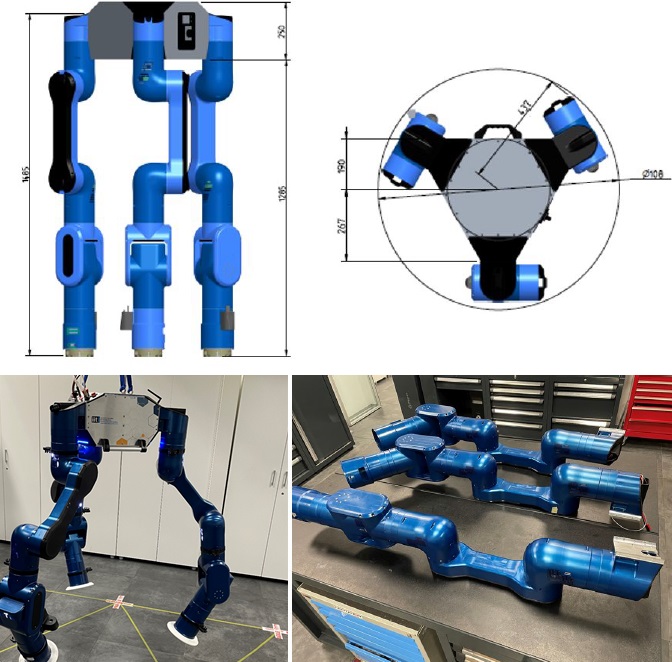}
    \caption{Final design and footprint of the MARM prototype.}
    \label{fig:mirror_proto}
    \vspace{-0.5cm}
\end{figure}

A prototype of the MARM robot has been manufactured following the design guidelines derived in the previous section.
The selected final MARM limb kinematics consists of 6-DoFs, which implement a 2-DoFs shoulder/hip complex, an elbow/knee joint, and a 3-DoFs wrist/ankle complex.
The three arms are mounted around a central body, in a triangular arrangement, as shown in Figure~\ref{fig:mirror_proto}.
The prototype weighs $94.4 \ kg$, with a pelvis mass of $25.1 \ kg$, and $23.1 \ kg$ per arm; with the legs fully extended, the robot is $1.47 \ m$ tall.  
\par
Based on the simulation studies performed during the preliminary design phase, two actuator sizes were identified based on the torque requirements results obtained from simulations.  
Based on this, IIT designed two sizes of actuation modules that were scaled/tuned to provide the torque demands identified for the different joints. The two actuator sizes were inherited from the foundation actuation family developed in previous projects at the HHCM lab~\cite{kashiri2019centauro}. The actuation units are realized based on two main assemblies: one consists of a frameless brushless DC motor, and the other includes a Harmonic Drive gearbox and the torque sensor unit. 
The actuators employ a 19-bit magnetic encoder measuring the absolute rotor position and serving the Field Oriented control implemented on the motor driver, and a 19-bit magnetic encoder for measuring the absolute link position. Using these two actuator sizes, 2-DoFs body modules were realized and interconnected to form the 6-DoFs arm kinematics derived from the simulation studies. 
A torque sensing load cell is mounted between the output link and the Harmonic Drive. 
%
\section{Preliminary MARM validation}\label{sec:experiments}

\par
We now introduce the preliminary results from tests performed on the real MARM platform prototype.
The MARM software architecture is based on the XBotCore~\cite{xbot} and CartesI/O~\cite{laurenzi2019cartesi} frameworks, used to implement and execute the testing motions.
%

%
%
\subsection{Gravity compensation testing}
\begin{figure}[htb!]
    \centering
    \includegraphics[width=1\columnwidth, trim={0cm 3cm 0cm 3cm}, clip=true]{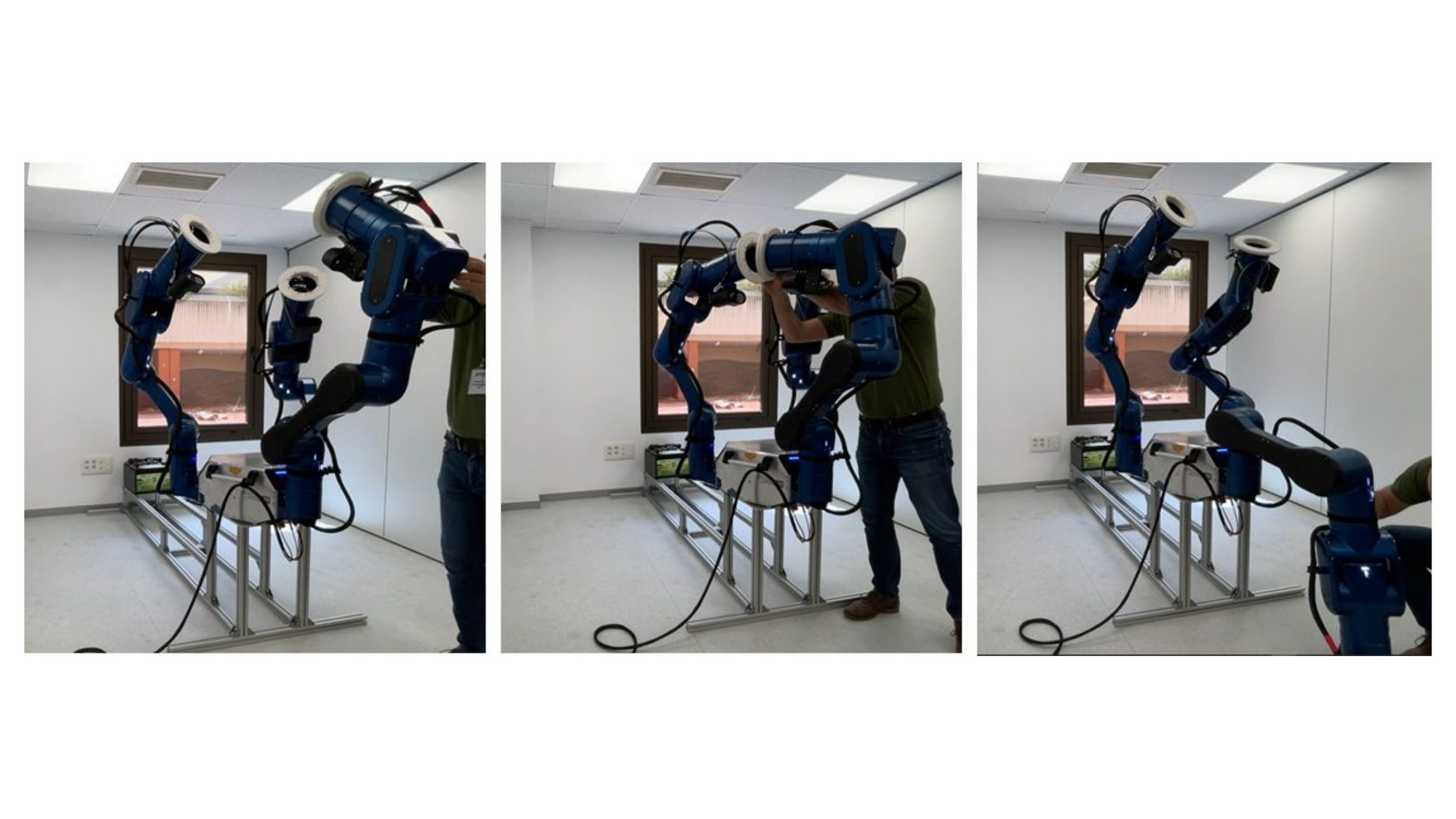}   
    \vspace{-0.5cm}
    \caption{A human operator freely manipulates the MARM arms under gravity compensation with joint impedance gains set to zero.}
    \label{fig:gravity_comp}
\end{figure}

\begin{figure}[htb!]
    \centering
    \vspace{-0.3cm}
    \includegraphics[width=1.\columnwidth, trim={2cm 0cm 2cm 0cm}, clip=true]{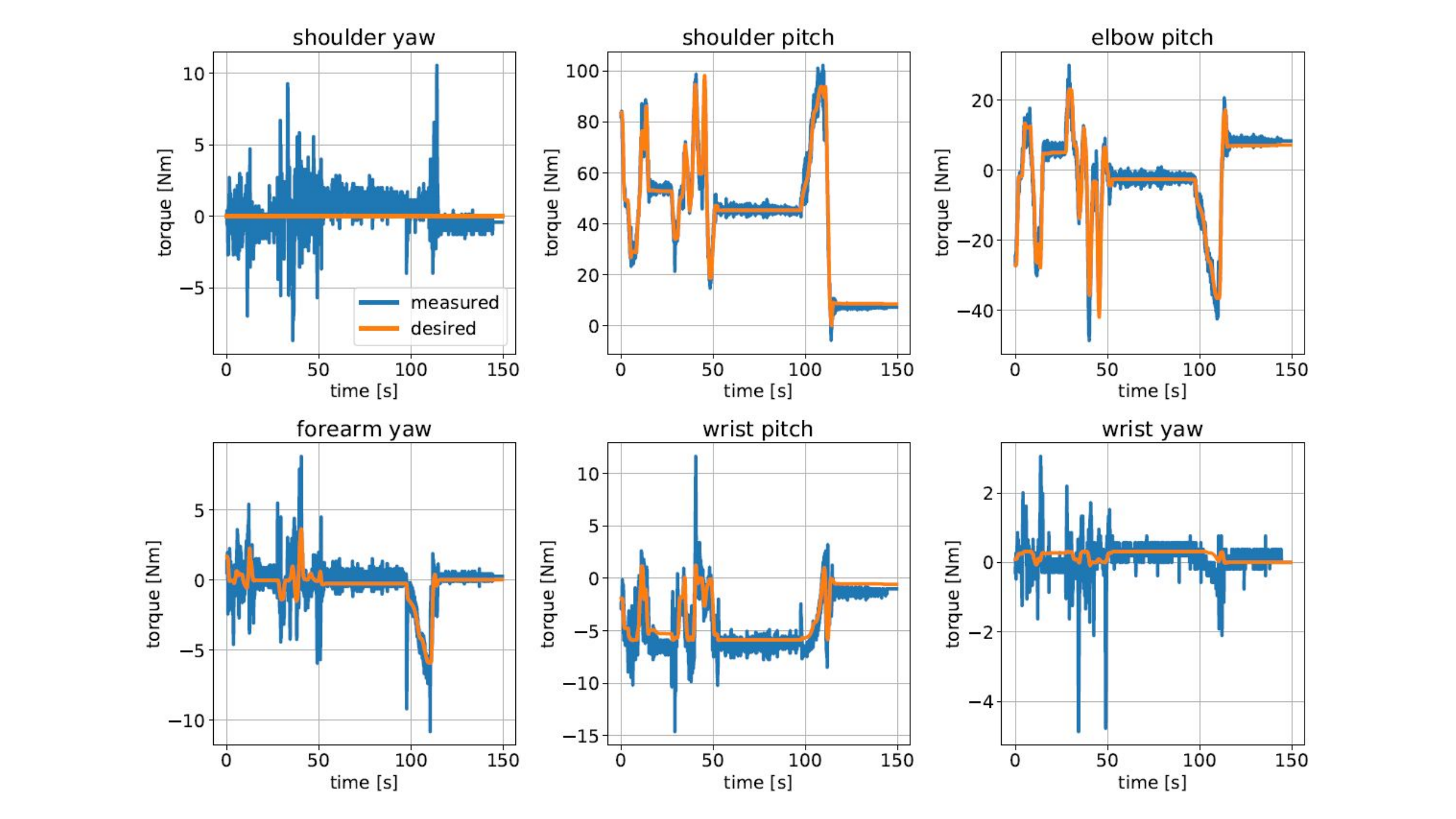}
    \vspace{-0.8cm}
    \caption{Joint torque tracking during gravity compensation.}
    \label{fig:gravity_comp_log}
    \vspace{-0.5cm}
\end{figure}

The purpose of this first assessment was to verify the synergistic consistency and accuracy of the MARM mass properties and URDF model in general, and the joint sensing and control performance. 
%
%
Starting from an initial configuration the feed-forward gravity torque was fed to the joint-level control boards as computed from the URDF model of the robot. 
Following the application of the feed-forward gravity compensation torque, the stiffness and damping gains of the joint impedance controller were reduced to zero leaving the three MARM arms free to move and supported only by the gravity compensation torque, see Figure~\ref{fig:gravity_comp}.  
The test confirmed the adequate synergistic performance of the robot model and joint torque sensing and control, Figure~\ref{fig:gravity_comp_log}, demonstrating minimal drift effects while a human operator was regulating freely the posture of the MARM arms.
\subsection{Assessment of Cartesian pose accuracy and repeatability}
The purpose of this test was to perform a first assessment of the Cartesian pose and motion accuracy of one of the MARM arms while executing a pick-and-place task.
The task was performed in a cyclic manner in order to evaluate the repeatability of the arm through a number of trials involving the reaching of the predefined target pose within the arm workspace.
\begin{figure}[htb!]
    \centering
    \includegraphics[width=0.95\columnwidth, trim={2cm 2cm 2cm 2.5cm}, clip=true]{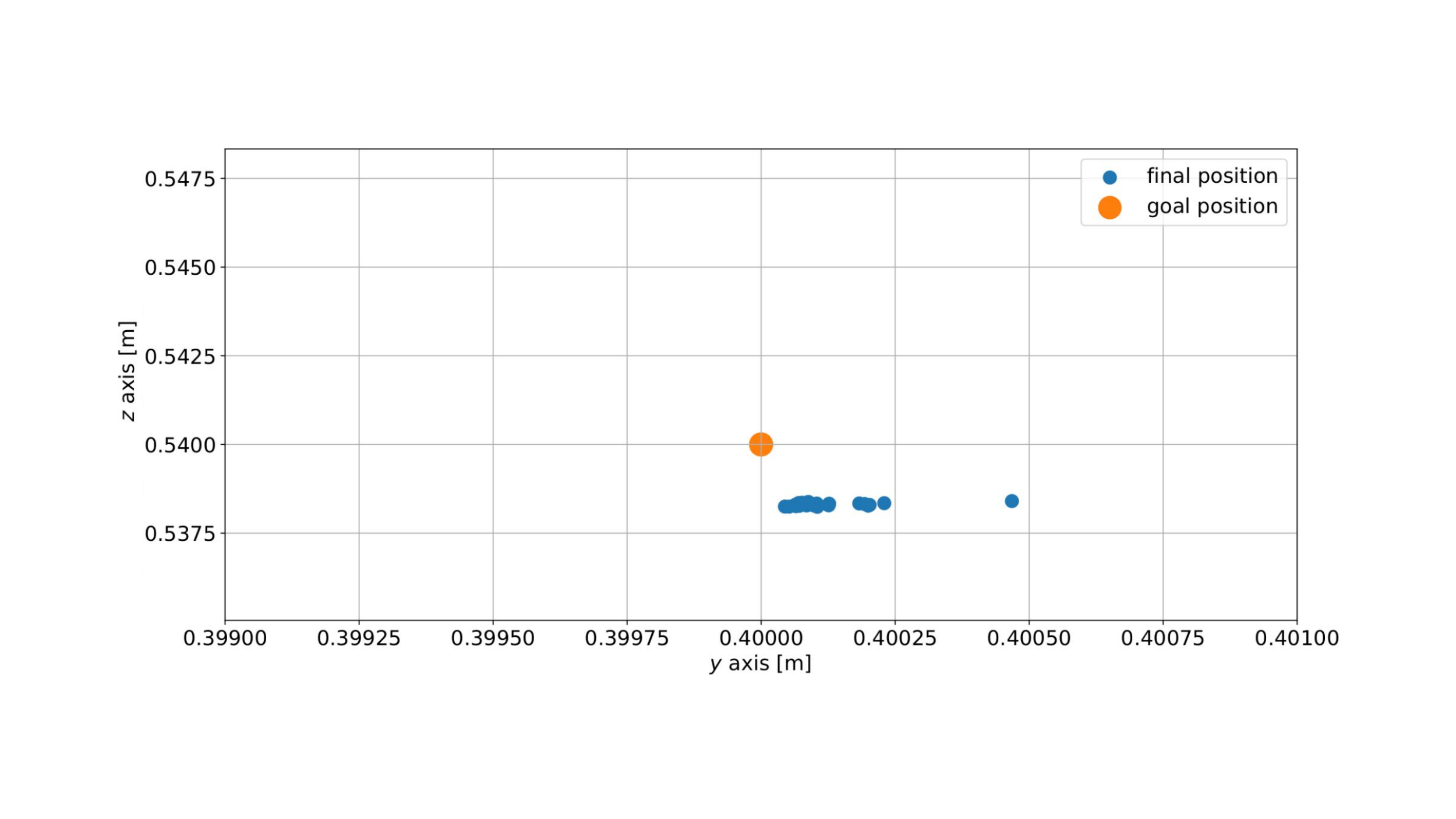}
    \vspace{-0.5cm}
    \caption{Cartesian poses reached performing the pick-and-place task with a single arm.}
    \label{fig:accuracy_repeatability}
\end{figure}
In Figure~\ref{fig:accuracy_repeatability} are reported the results in terms of the final Cartesian position achieved w.r.t. the goal, for the z and y axes.
It is possible to see that the MARM arm shows good performances both in terms of repeatability and precision.
%
All the trials can be located inside a circle of $0.12 \ [mm]$, with the maximum error on the y-axis under the $[mm]$, and for the z-axis under $2.5 \ [mm]$. 

\subsection{Cartesian Space testing in contact with the environment}
\begin{figure}[htb!]
    \centering
    \includegraphics[width=0.95\columnwidth, trim={0cm 3.5cm 0cm 3.4cm}, clip=true]{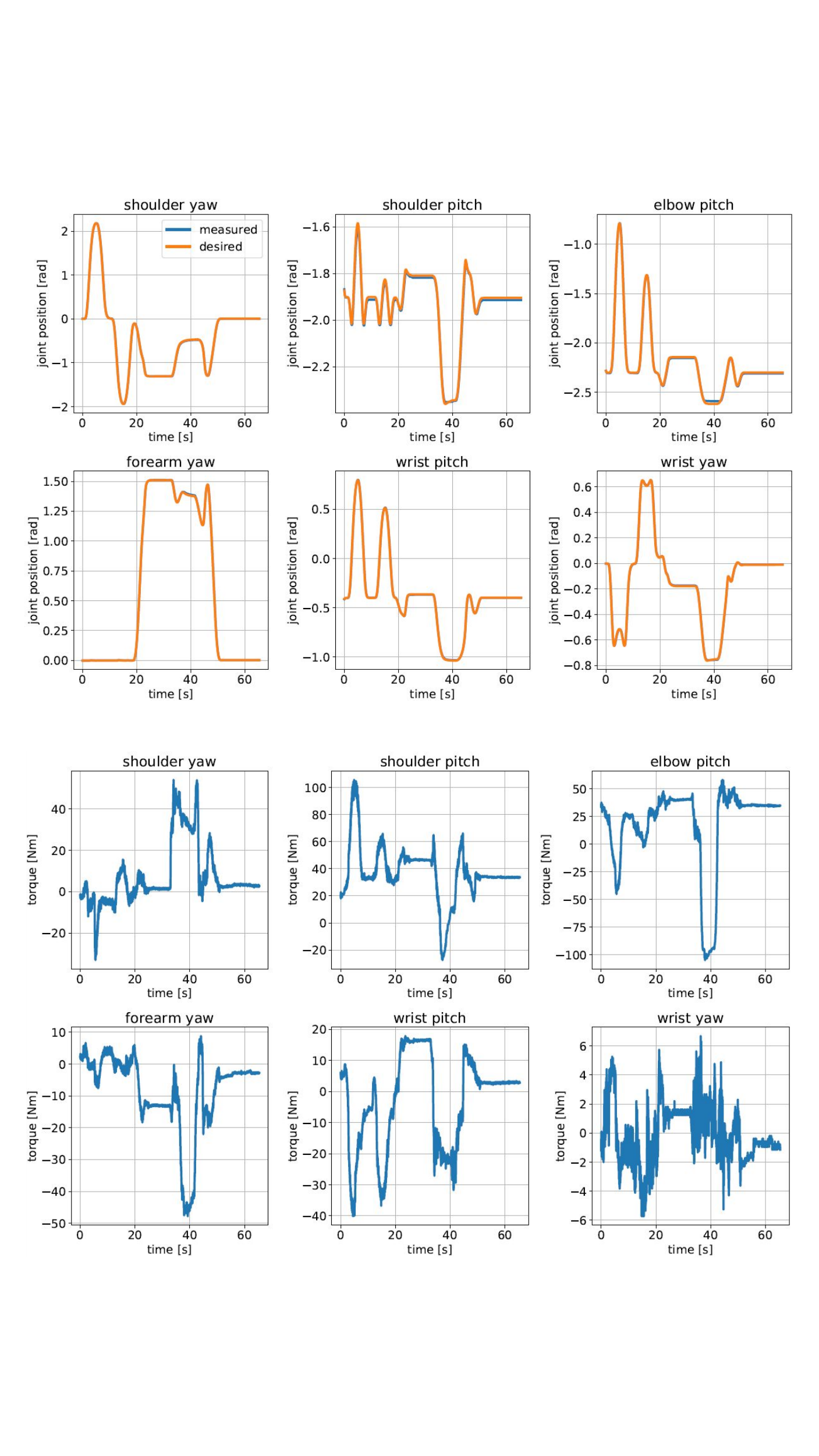}
    \caption{Joint tracking of one MARM limb and corresponding joint torques, while executing a set of Cartesian-space motions.}
    \label{fig:ground_tests}
\end{figure}

Further Cartesian-space tests have been carried out, where the robot is placed on the ground, and must therefore hold its own weight. 
We validated the final MARM performance by executing squatting motions, as well as balancing it on two feet, as depicted in Figure~\ref{fig:mirror_video_snaps}, and in the accompanying video, too. 
The attained tracking performance and joint torques are shown in Figure~\ref{fig:ground_tests}. 
%
%

\section{Conclusions}\label{sec:conclusions}
This work presented a codesign and validation pipeline applied to the study, design, and manufacturing of the MARM robot.
The proposed pipeline consists of numerical simulations based on sampling-based planning, trajectory optimization, and inverse dynamics/kinematics, to aid the selection of limb lengths, characterize motor sizing, and study the whole MARM kinematics according to the given task requirements, as specified by the MIRROR application by ESA.
We presented results from the proposed pipeline, as well as preliminary tests on the final hardware platform, demonstrating high-fidelity position and torque tracking, and advanced loco-manipulation capabilities.
Future works will comprise the deployment of the MARM robot in a use case scenario in which the robot will execute loco-manipulation tasks using the standard interfaces on its end-effectors to grasp, move and assemble tile modules with one of its arms while using the other two arms to grasp the floor tiles and locomote in the use case scenario infrastructure.  



\bibliographystyle{IEEEtran}	
\bibliography{biblio}

\begin{thebibliography}{10}
\providecommand{\url}[1]{#1}
\csname url@rmstyle\endcsname
\providecommand{\newblock}{\relax}
\providecommand{\bibinfo}[2]{#2}
\providecommand\BIBentrySTDinterwordspacing{\spaceskip=0pt\relax}
\providecommand\BIBentryALTinterwordstretchfactor{4}
\providecommand\BIBentryALTinterwordspacing{\spaceskip=\fontdimen2\font plus
\BIBentryALTinterwordstretchfactor\fontdimen3\font minus
  \fontdimen4\font\relax}
\providecommand\BIBforeignlanguage[2]{{%
\expandafter\ifx\csname l@#1\endcsname\relax
\typeout{** WARNING: IEEEtran.bst: No hyphenation pattern has been}%
\typeout{** loaded for the language `#1'. Using the pattern for}%
\typeout{** the default language instead.}%
\else
\language=\csname l@#1\endcsname
\fi
#2}}

\bibitem{ambrose2000robonaut}
R.~O. Ambrose, H.~Aldridge, R.~S. Askew, R.~R. Burridge, W.~Bluethmann,
  M.~Diftler, C.~Lovchik, D.~Magruder, and F.~Rehnmark, ``Robonaut: Nasa's
  space humanoid,'' \emph{IEEE Intelligent Systems and Their Applications},
  vol.~15, no.~4, pp. 57--63, 2000.

\bibitem{hiltz2001canadarm}
M.~Hiltz, C.~Rice, K.~Boyle, and R.~Allison, ``Canadarm: 20 years of mission
  success through adaptation,'' in \emph{International Symposium on Artificial
  Intelligence, Robotics and Automation}, 2001.

\bibitem{matsueda1991jemrms}
T.~Matsueda, K.~Kuraoka, K.~Goma, T.~Sumi, and R.~Okamura, ``Jemrms system
  design and development status,'' in \emph{IEEE National Telesystems
  Conference Proceedings}, 1991, pp. 391--395.

\bibitem{boumans1998european}
R.~Boumans and C.~Heemskerk, ``The european robotic arm for the international
  space station,'' \emph{Robotics and Autonomous systems}, vol.~23, no. 1-2,
  pp. 17--27, 1998.

\bibitem{vinals2020standard}
J.~Vinals, J.~Gala, and G.~Guerra, ``Standard interface for robotic
  manipulation (sirom): Src h2020 og5 final results-future upgrades and
  applications,'' in \emph{International Symposium on Artificial Intelligence,
  Robotics and Automation in Space (i-SAIRAS)}, 2020.

\bibitem{roa2017robotic}
M.~A. Roa~Garzon, K.~Nottensteiner, A.~Wedler, and G.~Grunwald, ``Robotic
  technologies for in-space assembly operations,'' in \emph{14th Symposium on
  Advanced Space Technologies in Robotics and Automation (ASTRA)}, 2017.

\bibitem{mukherjee2020robotically}
R.~Mukherjee, S.~Backus, T.~Setterfield, A.~Brinkman, G.~Agnes, E.~Sunada,
  J.~Kim, B.~Emanuel, R.~Smith, J.~Hyon, \emph{et~al.}, ``A robotically
  assembled and serviced science station for earth observations,'' in
  \emph{IEEE Aerospace Conference}, 2020, pp. 1--15.

\bibitem{lee2016architecture}
N.~N. Lee, J.~W. Burdick, P.~Backes, S.~Pellegrino, K.~Hogstrom, C.~Fuller,
  B.~Kennedy, J.~Kim, R.~Mukherjee, C.~Seubert, \emph{et~al.}, ``Architecture
  for in-space robotic assembly of a modular space telescope,'' \emph{Journal
  of Astronomical Telescopes, Instruments, and Systems}, vol.~2, no.~4, p.
  041207, 2016.

\bibitem{yoshida2006capture}
K.~Yoshida, D.~Dimitrov, and H.~Nakanishi, ``On the capture of tumbling
  satellite by a space robot,'' in \emph{IEEE/RSJ International Conference on
  Intelligent Robots and Systems}, 2006, pp. 4127--4132.

\bibitem{papadopoulos1993large}
E.~G. Papadopoulos, ``Large payload manipulation by space robots,'' in
  \emph{IEEE/RSJ International Conference on Intelligent Robots and Systems
  (IROS)}, vol.~3, 1993, pp. 2087--2094.

\bibitem{mcbryan2020comparison}
K.~McBryan, ``Comparison between stationary and crawling multi-arm robotics for
  in-space assembly,'' in \emph{IEEE/RSJ International Conference on
  Intelligent Robots and Systems (IROS)}, 2020, pp. 1849--1856.

\bibitem{rusconi2008dexarm}
A.~Rusconi, P.~Magnani, P.~Campo, R.~Chomicz, G.~Magnani, C.~Lambert,
  G.~Gruener, \emph{et~al.}, ``Dexarm engineering model development and
  testing,'' in \emph{10th ESA Workshop on Advanced Space Technologies for
  Robotics and Automation-ASTRA}, vol. 2008, 2008.

\bibitem{sirom_datasheet}
``Datasheet of the standard interface for robotic manipulation (sirom),''
  https://www.aeroespacial.sener/en/pdf-profile-project/standard-interface-for-robotic-manipulation-sirom.

\bibitem{sucan2012ompl}
I.~A. Sucan, M.~Moll, and L.~E. Kavraki, ``The open motion planning library,''
  \emph{IEEE Robotics \& Automation Magazine}, vol.~19, no.~4, pp. 72--82,
  2012.

\bibitem{nspg}
L.~Rossini, E.~M. Hoffman, A.~Laurenzi, and N.~G. Tsagarakis, ``Nspg: An
  efficient posture generator based on null-space alteration and kinetostatics
  constraints,'' \emph{Frontiers in Robotics and AI}, vol.~8, 2021.

\bibitem{OpenSot17}
E.~Mingo~Hoffman, A.~Rocchi, A.~Laurenzi, and N.~G. Tsagarakis, ``Robot control
  for dummies: Insights and examples using opensot,'' in \emph{IEEE/RAS
  International Conference on Humanoid Robots (Humanoids)}, Birmingham, UK,
  2017, pp. 736--741.

\bibitem{laurenzi2018balancing}
A.~Laurenzi, E.~M. Hoffman, M.~P. Polverini, and N.~G. Tsagarakis, ``Balancing
  control through post-optimization of contact forces,'' in \emph{IEEE-RAS
  International Conference on Humanoid Robots (Humanoids)}, 2018, pp. 320--326.

\bibitem{ruscelli2022horizon}
F.~Ruscelli, A.~Laurenzi, N.~G. Tsagarakis, and E.~Mingo~Hoffman, ``Horizon: A
  trajectory optimization framework for robotic systems,'' \emph{Frontiers in
  Robotics and AI}, vol.~9, 2022.

\bibitem{laurenzi2019cartesi}
A.~Laurenzi, E.~M. Hoffman, L.~Muratore, and N.~G. Tsagarakis, ``Cartesi/o: A
  ros based real-time capable cartesian control framework,'' in \emph{IEEE
  International Conference on Robotics and Automation (ICRA)}, 2019, pp.
  591--596.

\bibitem{kashiri2019centauro}
N.~Kashiri, L.~Baccelliere, L.~Muratore, A.~Laurenzi, Z.~Ren, E.~M. Hoffman,
  M.~Kamedula, G.~F. Rigano, J.~Malzahn, S.~Cordasco, \emph{et~al.},
  ``Centauro: A hybrid locomotion and high power resilient manipulation
  platform,'' \emph{IEEE Robotics and Automation Letters}, vol.~4, no.~2, pp.
  1595--1602, 2019.

\bibitem{xbot}
L.~Muratore, A.~Laurenzi, E.~Mingo~Hoffman, and N.~G. Tsagarakis, ``The xbot
  real-time software framework for robotics: From the developer to the user
  perspective,'' \emph{IEEE Robotics \& Automation Magazine}, vol.~27, no.~3,
  pp. 133--143, 2020.

\end{thebibliography}

\end{document}